\documentclass[journal,twocolumn]{IEEEtran}
\usepackage{epsfig}
\usepackage{graphicx}
\usepackage{amsmath}
\usepackage{amssymb,subfigure,cases}
\usepackage{setspace}
\usepackage{cite}

\usepackage{color,hyperref}
\usepackage{algorithm}
\usepackage{algorithmic}
\usepackage{multirow}
\usepackage{url}
\usepackage{booktabs}
\usepackage{stfloats}
\usepackage{soul}

\ifCLASSINFOpdf

\else

\fi

\definecolor{darkblue}{rgb}{0.0,0.0,1.0}
\hypersetup{colorlinks,breaklinks,
            linkcolor=darkblue,urlcolor=darkblue,
            anchorcolor=darkblue,citecolor=darkblue}

\begin{document}
\setulcolor{red}

\title{Anomaly Detection in Traffic Scenes via Spatial-aware Motion Reconstruction}

\author{~Yuan~Yuan,\IEEEmembership{~Senior Member,~IEEE}, Dong~Wang, and Qi~Wang$^{*}$,\IEEEmembership{~Senior Member,~IEEE}
\thanks{This work is supported by National Natural Science Foundation of China under Grant 61379094, Natural Science Foundation Research Project of Shaanxi Province under Grant 2015JM6264 and Fundamental Research Funds for the Central Universities under Grant 3102015BJ(II)JJZ01.

The authors are with School of Computer Science and Center for OPTical IMagery Analysis and Learning (OPTIMAL),
Northwestern Polytechnical University, Xi'an 710072, Shaanxi, P. R. China.

Qi Wang is the corresponding author (e-mail: crabwq@nwpu.edu.cn).}
}

\markboth{{IEEE} Transactions on Intelligent Transportation Systems}%
{Shell \MakeLowercase{\textit{et al.}}: Bare Demo of IEEEtran.cls for Journals}

\maketitle

\begin{abstract}
Anomaly detection from a driver's perspective when driving is important to autonomous vehicles. As a part of Advanced Driver Assistance Systems (ADAS), it can remind the driver about dangers timely. Compared with traditional studied scenes such as the university campus and market surveillance videos, it is difficult to detect abnormal event from a driver's perspective due to camera waggle, abidingly moving background, drastic change of vehicle velocity, etc. To tackle these specific problems, this paper proposes a spatial localization constrained sparse coding  approach for anomaly  detection in traffic scenes, which firstly measures the abnormality of motion orientation and magnitude respectively and then fuses these two aspects to obtain a robust detection result. The main contributions are threefold: 1) This work describes the motion orientation and magnitude of the object respectively in a new way, which is demonstrated to be better than the traditional motion descriptors. 2) The spatial localization of object is taken into account of the sparse reconstruction framework, which utilizes the scene's structural information and outperforms the conventional sparse coding methods. 3) Results of motion orientation and magnitude are adaptively weighted and fused by a Bayesian model, which makes the proposed method more robust and handle more kinds of abnormal events. The efficiency and effectiveness of the proposed method are validated by testing on nine difficult video sequences captured by ourselves. Observed from the experimental results, the proposed method is more effective and efficient than the popular competitors, and yields a higher performance.
\end{abstract}


\IEEEpeerreviewmaketitle

\section{Introduction}
\label{Introduction}
\begin{figure}
\centering
\includegraphics[width=.48\textwidth]{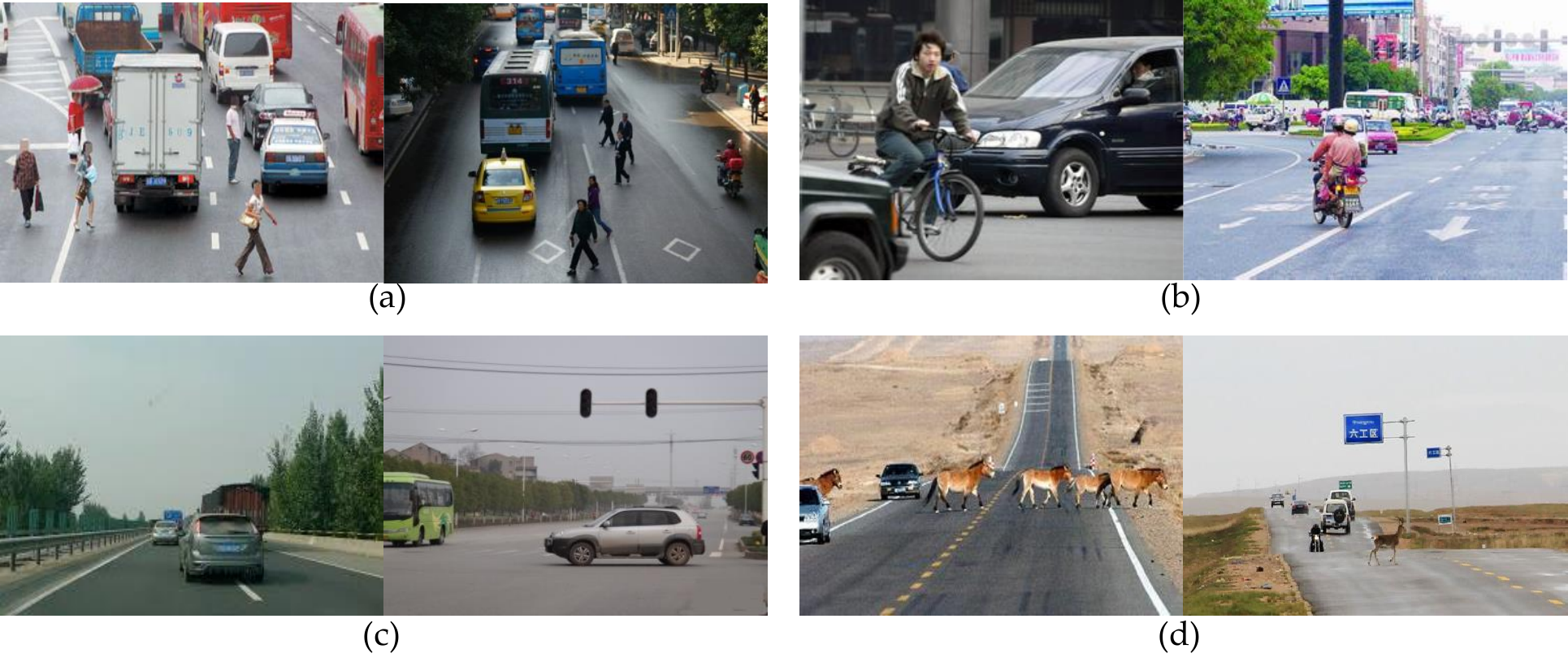}
\caption{Typical examples of anomaly in traffic scenes. (a) Pedestrian crossing the road; (b) Cyclists and motorcyclists on the road; (c) Vehicle overtaking and (d) sudden appearance of animals. It is noticed that the abnormal objects have a different compared to its neighboring object.}\label{Fig-Problem}
\end{figure}

\IEEEPARstart{T}{here} are many potential dangers when driving, such as unsafe driver behavior, sudden pedestrian crossing, and vehicle overtaking. Fig. \ref{Fig-Problem} shows some typical exemplars having potential dangers. Since the driver's attention can't focus in every second and notice all dangers, many traffic accidents occur every day. Therefore, it is necessary to auto-detecting potential dangers from a driver's perspective, and a surge of interests has been motivated in computer vision community. But it is almost impossible to design a system that can detect faultlessly all kinds of abnormal event, because the anomaly definition might be distinctive in different situations. Therefore, many researchers simplify the problem by focusing on specific objects and events, such as pedestrians, vehicles and crossing behaviors.

To tackle the above simplified problem, training object detectors is a straightforward method. To name only a few, Xu \emph{et al.} \cite{DBLP:journals/tsmc/XuXLHCL12} focus on detecting the sudden crossing pedestrians when driving, and learn a pedestrian detector to detect crossing pedestrians as early as possible. Sivaraman and Trivedi \cite{6338886} propose a part-based vehicle detector to detect cars when driving. Moreover, to improving accuracy of the detector, Garcia \emph{et al.} \cite{DBLP:conf/fusion/GarciaEAHL11, 6082971} fuse vision-based pedestrian detection results and laser data to estimate the frontal pedestrian. Apart from these traditional methods, over recent years, the landscape of computer vision has been drastically altered and pushed forward through the adoption of deep learning, especially the Convolutional Neural Network (CNN) \cite{lecun1998gradient}. The CNN-based object detectors achieve state-of-the-art results in almost all object detection benchmarks. As an example, Region-based CNN \cite{girshick2014rich} achieves excellent object detection accuracy by using deep ConvNet to classify object proposals. Based on the similar framework, there are quite a few works to speed up R-CNN such as Spatial Pyramid Pooling networks (SPPnets) \cite{he2015spatial} and Fast R-CNN \cite{girshick2015fast}. Though the CNN-based object detection method is outstanding in static image, the trained models only capture appearance information and cannot be used to recognize specific actions immediately.

There is another clue to classification of different behaviors by contrast with static image, $i.e.$, object motion information. A slice of papers investigate for action detections in this direction. Early work by Alonso \emph{et al.} \cite{DBLP:journals/tvt/AlonsoVRM08} detects the overtaking cars in reference to the motion orientation of vehicles, which is obtained by calculating the optical flow of every frame. Along similar line, Kohler \emph{et al.} \cite{6338797} propose a Motion Contour image based on HOG-like descriptor (MCHOG) in combination with a SVM learning algorithm that decides within the initial step if a pedestrian at the curb will enter the traffic lane. Aside from these motion flow based methods, object trajectory is another technique for describing object motion information. As an example, Bonnin \emph{et al.} \cite{6957720} propose a generic model to predict pedestrians crossing behavior in inner-city, which predicts the pedestrain's motion orientation by tracking for a while. However, because object tracking is not credible all the time in fickle scenes, the object trajectory is misleading to object localization. This limitation makes it unfavorable in traffic scene. Besides, the tracking technique usually needs the target to be detected as an initial step, which makes the method also object-related.

A desirable property of a system which is able to identify threats when driving is to disentangle specific object classes. The detector-based and tracking-based methods invariably pour attention into quite a few object. Consequently, this work resorts to the motion flow based method. However, in order to make motion flow based method feasible, there are several difficulties should be considered carefully. First, since the camera is mounted on the moving vehicle, it is almost shaking all the time and the captured video is usually blurred. This makes the estimated motion information noisy and unstable. Second, in contrast to the static camera, the background of scene is all moving due to its relative movement to the camera, which makes the motion patterns of the scene very complex. Additionally, the ever-changing background makes the influence of background more serious. Third, there is some drastic variation of vehicle velocity, aggravating the difference of relative movements between objects. Due to dynamic uncertainty, the same behaviors such as sudden vehicle crossing , may show totally different motion patterns with different vehicle velocities.

In order to tackle the above problems, this work calculates two histograms to represent motion magnitude and orientation respectively, which makes a more comprehensive description of local motion pattern, and the separate descriptors have a clearer expression of motion patterns resulting in resistance of motion noise. Additionally, two anomaly maps are generated by spatial-aware reconstruction, which can alleviate the influence of dynamic background via spatial constraint. Finally, a Bayesian integration model is employed to fuse previously obtained anomaly maps to calculate the final anomaly map, which is robust to the drastic changes of vehicle velocity. Based on the obtained final anomaly map, the abnormal objects can be located.

The reminder of this paper is organized as follows: Section \ref{Related Work} reviews previous work on anomaly detection in computer vision. The main steps and contributions of the proposed method are clarified briefly in Section \ref{Overview}. Section \ref{motion segmentation} describes the strategy for motion region segmentation. Section \ref{anomaly detection} proposes the anomaly detection and localization using sparse reconstruction. The Bayesian-based integration method is elaborated in Section \ref{Bayesian integration} and experiments and discussions are given Section \ref{experments and discussion}. The conclusion is finally summarized in Section \ref{conclusion}.
\section{Related Work}
\label{Related Work}
The proposed framework in this paper bears some resemblance to region of interest (ROI) generation and selection methods, and measures the degree of anomaly via sparse reconstruction cost in conjunction with the integration of two motion clues that is inspired by multi-saliency evaluation. Hence the literature review for this work begins from these three aspects.

In the realm of the relative works for ROI generation and selection, there are several efforts \cite{Uijlings2013, zitnick2014edge, cheng2014bing, girshick2016region} creating a relatively small set of candidate ROIs that cover the objects in the image. The ``selective search" algorithm of van de Sande \emph{et al.} \cite{Uijlings2013} computes hierarchical segmentations of superpixel \cite{felzenszwalb2004efficient} and places bounding boxes around them. EdgeBoxes \cite{zitnick2014edge} outputs high-quality rectangular (box) proposals quickly, which are selected readily with a simple box objectness score computed from the contours wholly enclosed in a candidate bounding box. Additionally, BING \cite{cheng2014bing} trains a two stages cascaded SVM \cite{zhang2011proposal} to measure generic objectness, and then produces a samll set of candidate object windows. Finally, recent R-CNN \cite{girshick2016region} applies high-capacity convolutional networks to bottom-up region proposals in order to localize and segment objects, and gives more than a 50\% relative improvement on PASCAL VOC. Our approach is inspired by the success of these ROI selection methods, and the difference is we filtrate ROIs according to measuring abnormality, rather than objectness.

There are quite a few alternatives to model the degree of anomaly, such as mixture of probabilistic principal component analysis (MPPCA) model \cite{DBLP:journals/pami/LiMV14}, social force model \cite{DBLP:conf/cvpr/MehranOS09}, sparse basis \cite{DBLP:journals/pr/CongYL13, lu2013abnormal, zhao2011online, mo2014adaptive}, etc. However, based on the sparsity of unusual events, more and more sparsity based methods have emerged in this field recently. Cong \emph{et al.} \cite{DBLP:journals/pr/CongYL13} calculate a multiscale histogram of optical flow to represent the local motion patterns for image sequences. Whether a testing sample is abnormal or not is determined by its sparse reconstruction cost, through a weighted linear reconstruction of the over-complete normal basis set. Zhao \emph{et al.} \cite{zhao2011online} propose a fully unsupervised dynamic sparse coding approach for detecting unusual events in videos based on online sparse reconstructibility of query signals from an automatically learned event dictionary, which forms a sparse coding bases. Moreover, recent research has observed and validated that locality is more essential than sparsity \cite{5540018, yu2009nonlinear, liu2011defense}. The locality-constrained linear coding (LLC) [38] is a great advance in this aspect, which applies locality constraint to select similar basis of local image descriptors. Inspired by this work, we measure abnormality by spatial locality-constrained sparse reconstruction.

For obtaining robust and superior results, integration of multiple clues or factors is usually adopted in computer vision and machine learning community. Because of close similarity between anomaly map and saliency map, we review some work about multi-saliency fusion here. The straightforward and most intuitive scheme is linear fusion. Evangelopoulos \emph{et al.} \cite{evangelopoulos2013multimodal} apply this framework to fuse aural, visual and textual saliency. For more elaborate fusion, a Support Vector Machine is trained and used to predict the quality of each saliency map in \cite{yang2015discovering}, and then saliency maps are fused linearly using the quality measure of each map. Besides, Xie \emph{et al.} \cite{DBLP:journals/tip/XieLY13} merge low and mid level visual saliency within the Bayesian framework, which generates more discriminative saliency map. Furthermore, the Bayesian integration method is also employed in \cite{DBLP:conf/iccv/LiLZRY13} and performs better than the conventional integration strategy.

\section{Overview}
\label{Overview}
In this paper, an effective anomaly detection method for traffic scenes is designed, which is robust to the change of the camera movement. And the components and contributions of this method is illuminated schematically in this section.

\begin{figure*}
\centering
\includegraphics[width=0.95\textwidth]{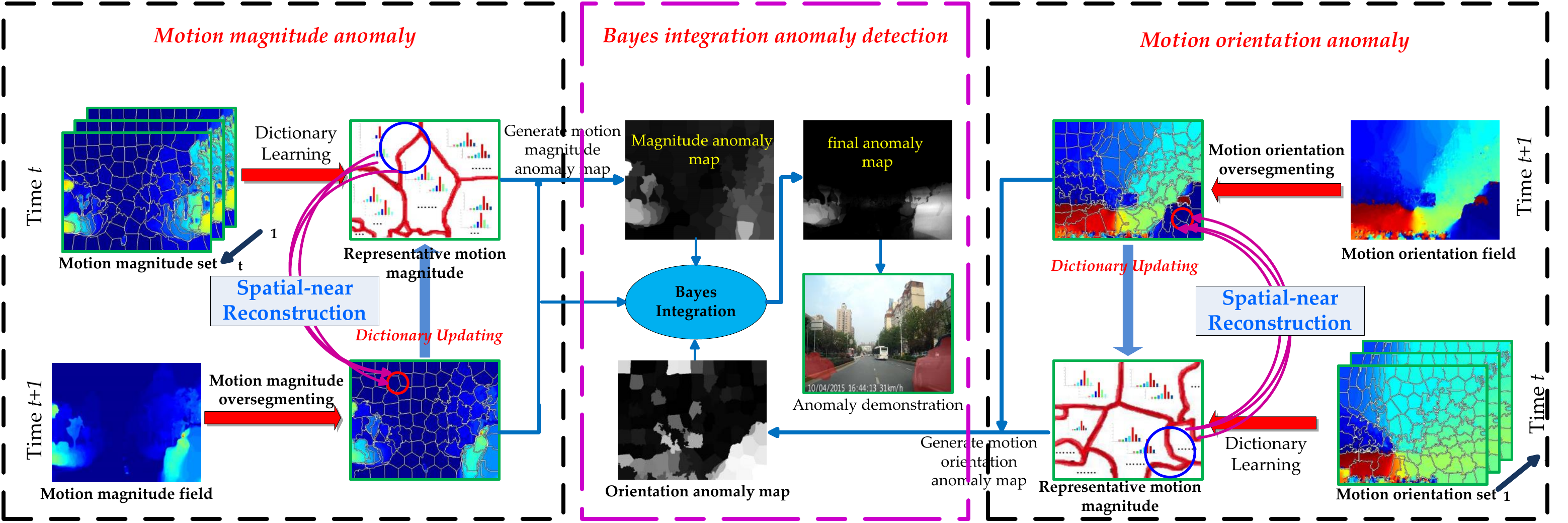}
\caption{The pipeline of the proposed method. First, with the obtained motion estimation which is computed by a state-of-the-art dense flow method \cite{liu2009beyond}, the optical flow field is separated into two motion fields, $i.e.$, motion orientation field and motion magnitude field. Then SLIC \cite{achanta2012slic} superpixel segmentation is utilized to over-segment each motion field into superpixels. Second, with the superpixel motion features of both motion fields, this work learns two dictionaries respectively for the motion orientation and magnitude and updates the learned dictionaries to adapt to dynamic scenes. The newly observed superpixel motion feature is reconstructed by its top $K$ nearest elements of the corresponding dictionary. The superpixel motion features with large reconstruction error are not used to update the corresponding dictionary. Third, in order to give a more robust anomaly estimation, this work integrates the obtained two anomaly map based on Bayes model, which makes use of the complementarity between motion orientation and magnitude. In the end, the detected anomaly regions are superimposed on the original color image.}\label{Fig-Method}
\end{figure*}

\subsection{Components of the Proposed method}
The main components are illustrated in Fig. \ref{Fig-Method}, with a detailed description as follows.
\subsubsection{Complementary motion description}
Given a video sequence, this work calculates the optical flow field of each frame, which represents the motion characteristics of each pixel as a two-dimensional vector. With the obtained optical flow, the motion orientation and magnitude of each pixel is calculated and gathered together to form the motion orientation filed (MOF) and motion magnitude field (MMF) respectively. Since different parts of an object may have similar motion characteristics, the superpixel technique is employed to over-segment the obtained MOF and MMF, which can separate different objects well by preserving coherence of local motion patterns. With the segmented results, this work calculates a histogram for every superpixel to represent its motion orientation and magnitude. Because this technique takes these two aspects into consideration, the proposed method can detect motion orientation and magnitude anomaly simultaneously.
\subsubsection{Abnormality measurement via spatial-aware reconstruction}
With the obtained motion orientation and magnitude histogram, this work detects the motion orientation and magnitude anomaly simultaneously via a dictionary-based method. To be specific, this work learns two normal dictionaries respectively for motion orientation and magnitude description by an incremental learning method, which finds the representative samples (histogram of motion orientation or magnitude) in the normal motion pattern set. And then we construct the dictionary via taking them as the bases of the learned dictionary. For the reason that the location of motion feature (i.e., the spatial location of the corresponding superpixel) is essential to anomaly detection in traffic scene, this work reconstructs the newly observed motion feature over the spatial-near subset of the learned dictionary, which is inspired by the locality-constrained linear coding (LLC) \cite{5540018} method in image classification. Besides, in order to measure the difference of motion features more reasonably, the earth mover's distance (EMD) \cite{4135678} is employed instead of traditional $\chi_2$ distance. According to the reconstruction cost, two anomaly maps are generated and indicate abnormality of motion orientation and magnitude respectively.
\subsubsection{Bayesian-based integration of anomaly detection}
As mentioned above, this work measures the abnormality of motion orientation and motion magnitude simultaneously, and the behind idea is that some abnormal behaviors show a different motion orientation but some is motion magnitude, which is mainly caused by drastic changes of vehicle velocity. In order to tackle this problem, we integrate the two anomaly maps based on a Bayesian integration model via adaptive weights, which can make use of the complementarity between these two maps and obtain a robust detection result.
\subsection{Contributions}
In this work, we tackle the anomaly detection in traffic scenes via measuring the change of motion orientation and motion magnitude simultaneously and integrating these two complementarity aspects together to relieve the mobile camera problem. Additionally, the proposed method does not need any extra training video to pre-learn a off-line model. The main contributions of this paper are described as follows.
\begin{enumerate}
\item[1) ]
Explore different effects of motion orientation and magnitude on anomaly detection respectively and model them using a histogram-based method, which is suitable and reasonable to describe motion patterns in traffic video with mobile camera. Compared with the application scenes of traditional anomaly methods, which usually contain several simple motion patterns because of the static camera, the motion patterns in our scenery are more complex and noisy. The reason behind this is that the camera is shaking when driving and not in a constant velocity. Therefore, in order to increase the discriminability of the descriptor, this work calculates two histograms to represent motion orientation and magnitude respectively, which can eliminate the noise more easily.
\item[2) ]
Propose a spatial-aware spare reconstruction method to measure the abnormity of local motion patterns, which is achieved by reconstructing the newly observed motion pattern over its spatial-near dictionary elements. In previous literatures on anomaly detection, sparse reconstruction is utilized in some efforts, but they almost do not take the spatial information into consideration for the simplicity of application scenes. On the contrary, since the motion patterns in traffic video usually have a strong relationship with its spatial location, we reconstruct them with its spatial-near dictionary elements. It can eliminate the dynamic background influence and outperform the traditional sparse reconstruction method.
\item[3) ]
Introduce a Bayesian integration method to adaptively fuse the anomaly results from motion orientation and magnitude. Since the obtained two results usually have different efforts in different scenarios and are complementary to each other, this work integrates these preliminary results into a final detection result. Compared with the conventional integration strategy, such as addition and multiplication, which usually predetermine the integration weights, the employed Bayesian-based method takes the video content into consideration and allocate integration weights adaptively. Therefore, the Bayesian integration method can better reflect the video content and handle drastic changes of vehicle velocity.
\end{enumerate}

\begin{figure}
\centering
\includegraphics[width=.48\textwidth]{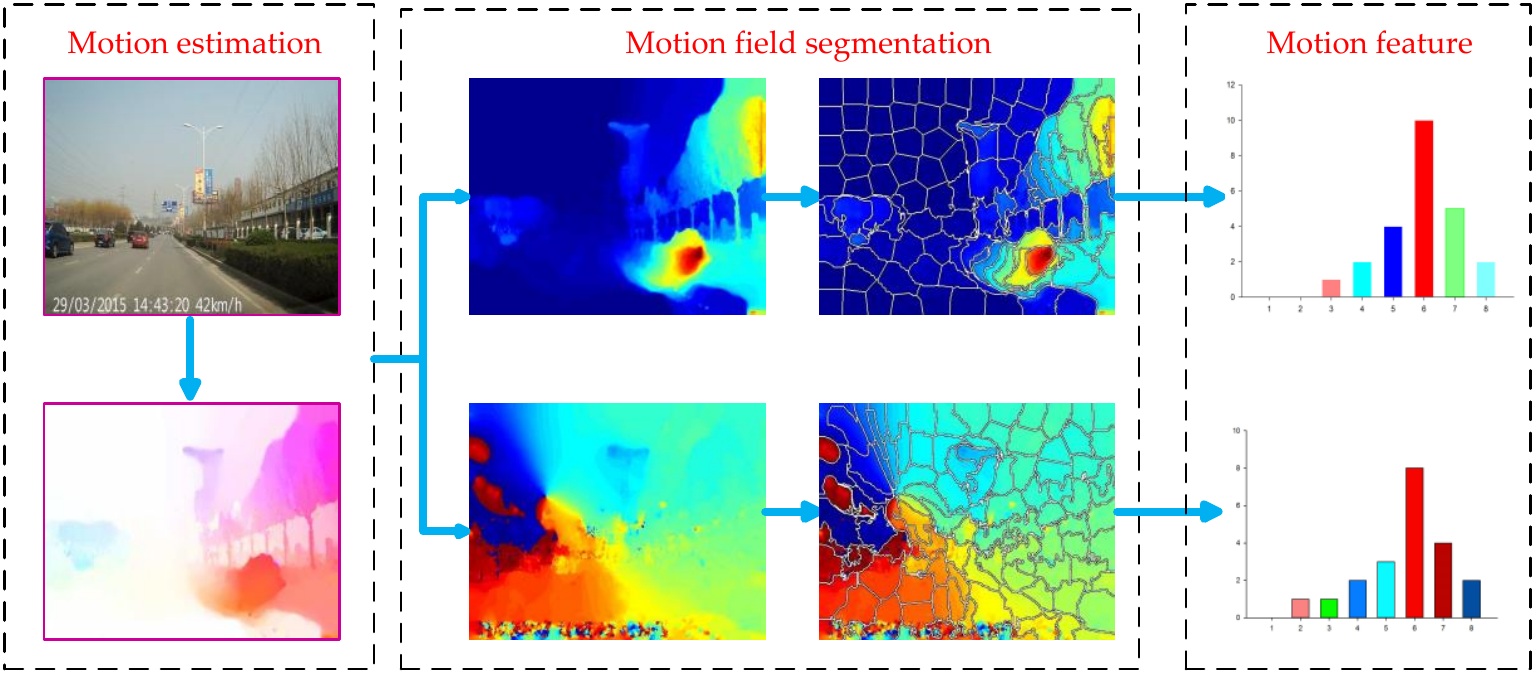}
\caption{The flowchart for motion feature extraction.}\label{Fig-MotionSeg}
\end{figure}

\section{complementary motion description}
\label{motion segmentation}
As we all know, traffic scenes are typically crowded. There is much occlusion when you driving on a road, which makes the trajectory-based approaches infeasible in this situation. As a main alternative, motion-based approaches show a promising result  for anomaly detection. Therefore, our proposed approach makes use of motion information instead of tracking individuals in the scene. For describing motion patterns effectively,  optical flow method\cite{liu2009beyond} is employed.

\subsection{Superpixel motion segmentation}
Since motion orientation and magnitude of different parts that belong to one object are homologous, the superpixel technique, which has a powerful ability for preserving image local coherence, is employed to segment different motion regions. To be specific, the optical flow field is separated into motion orientation field and motion magnitude field and the superpixels are obtained from   both   fields respectively. In detail, as illustrated in Fig. \ref{Fig-MotionSeg}, these two motion fields are converted into two gray-scale images, and then SLIC method \cite{achanta2012slic} is employed to over-segment these two "images" because of its low computational cost and high performance.

\subsection{Complementary motion representation}
\begin{figure}
\centering
\includegraphics[width=.5\textwidth]{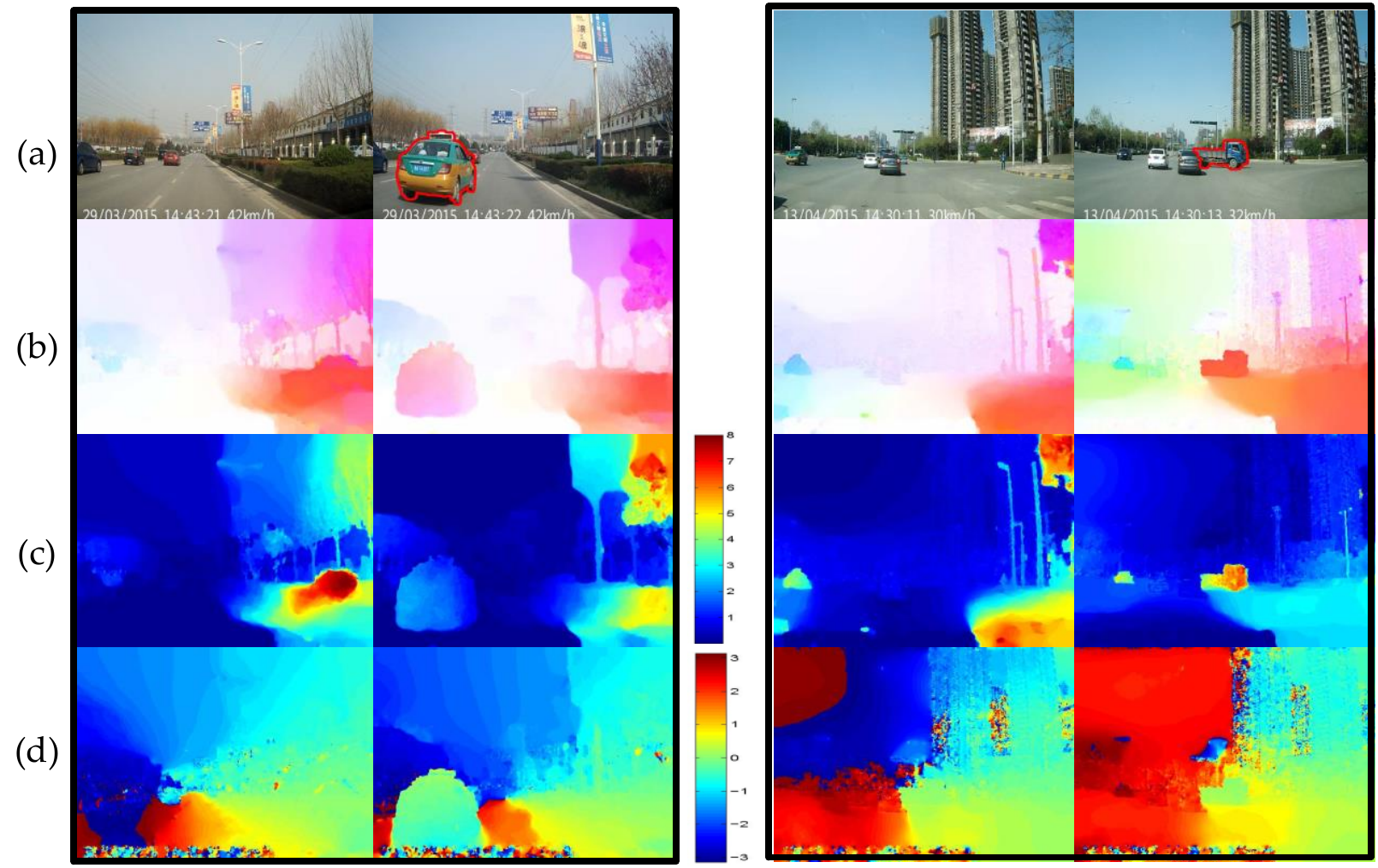}
\caption{Two abnormal events in traffic scenes, which show the complementarity of motion orientation and magnitude. (a) Original color image. The red circles denote the abnormal objects. (b) Optical flow field. It represents the motion information of every pixel. (c) Motion magnitude field. Different colors represent  different motion magnitudes. (d) Motion orientation field. Different colors represent  different motion orientations. It is obvious that   motion orientation is more discriminative than motion magnitude in the first scenario, and motion magnitude is more important in the second scenario.}\label{Fig-motionShow}
\end{figure}

 With the obtained superpixels, a histogram-based descriptor is calculated to represent motion information. The traditional histogram of orientated optical flow (THOOF) \cite{chaudhry2009histograms} sums the magnitude of optical flow according to its orientation followed by a normalization operation, which loses the motion magnitude clue\cite{Yang2011Sparse}. Considering that the anomaly definition in traffic scenes is usually different as illustrated in Fig. \ref{Fig-motionShow}, these two  factors are measured simultaneously and integrated to detect anomaly efficiently.

Suppose the motion orientation field  image is over-segmented into $N$ superpixels. For \emph{i-th} superpixel $sp_{oi}, i= 1,...,N$, its motion feature is denoted as $y_{oi} \in {R^{1 \times d}}$, where $d$ indicates the histogram dimension. In addition, the spatial location of \emph{i-th} superpixel centroid is represented by a two-dimensional coordinate $z_{oi} \in {R^2}$. And the whole set of these superpixels are denoted as $Y_o$ and $Z_o$.  Similarly, the \emph{i-th} superpixel of motion magnitude field is denoted as $sp_{mi}$, and its motion feature and spatial location are denoted as $y_{mi}$ and $z_{mi}$, whose whole sets are denoted as $Y_m$ and $Z_m$ respectively .

The distance measurement  between histograms  is essential   in the histogram-based method. Since the extracted optical flows are inevitably noisy and uncertainty, we adopt the earth mover's distance (EMD) as histogram distance function, which is a well-known robust metric in case of noisy histogram comparison. Specifically, the EMD between histogram $P$ and $Q$ is denoted as:
\begin{equation}
\label{Eq-EMD-distance}
\begin{array}{l}
EMD(P,Q) = {\kern 1pt} \,\mathop {\min }\limits_{{f_{ij}} > 0} \;\sum\limits_{i = 1}^d {\sum\limits_{j = 1}^d {{f_{ij}}{Dis_{ij}}} }, \\
s.t.\;\quad \sum\limits_{i = 1}^d {{f_{ij}}}  \le {P_j},\quad \sum\limits_{j = 1}^d {{f_{ij}}}  \le {Q_i}
\end{array}
\end{equation}
where $f_{ij}$ denotes a flow from bin $P(i)$ to $Q(j)$, and $Dis_{ij}$ is their ground distance. In general, the ground distance $Dis_{ij}$ can be any distance measurement, such as $l_1$ and $l_2$. For simplification, $l_1$ distance is employed in this work, which is:
\begin{equation}
\label{Eq-L1-distance}
{Dis_{ij}} = \left| {i - j} \right|.
\end{equation}
For reducing computation cost, we utilizes the EMD-$l_1$ instead of original EMD with $l_1$ ground distance. The equivalence of these two distances was verified in \cite{4135678} and the EMD-$l_1$ has a lower time complexity.

\section{Abnormality measurement via spatial-aware reconstruction}
\label{anomaly detection}
With the separated motion fields, the following task is to detect anomaly by measuring motion  inconsistency. This paper formulates the problem of anomaly detection as the reconstruction of the newly observed local motion pattern by the historically collected normal motion patterns.  Inspired by the Locality-constrained Linear Coding (LLC), more emphasis is laid on the spatial priors of the dictionary element. Moreover, the spatial prior is essential to alleviate the influence of the background motion patterns. Therefore, the reconstruction error of each superpixel's motion pattern is calculated by  its spatially near elements in the dictionary, which is learned \cite{DBLP:conf/cvpr/ElhamifarSV12} via finding the representative normal motion patterns. In the following, the dictionary learning method is introduced firstly, and then the estimation approach of anomaly via spatial neighbor reconstruction is presented.

\subsection{Dictionary learning via finding the representative motion patterns}
\label{Dictionary learning}
 For the  camera captured video in traffic scene,  the motion pattern has a strong spatial dependency.   Certain  motion patterns usually arise at specific spatial locations and different regions are prone to show different motion prototypes. In order to describe them, we find a few representative motion  patterns and retain its corresponding spatial localization.

To be specific, we measure the superpixel motion pattern's ability to reconstruct other normal motion patterns according to corresponding reconstruction coefficient, which is obtained by minimizing the reconstruction error of the all superpixel motion patterns. Similar to sparse reconstruction problem, the above optimization problem can be formalized as:
\begin{equation}
\label{Eq-DictLearning-1}
\mathop {\min }\limits_C \;\frac{1}{2}\left\| {Y - YC} \right\|_F^2\quad\;\;\;s.t.\;\;\;{\left\| C \right\|_{1,2}} < \varepsilon ,\;\;diag(C) = {\bf{0}},
\end{equation}
where $Y \in {R^{c \times N}}$ denotes the normal superpixels' motion patterns, $c$  the dimensionality of motion feature and $N$ the number of normal superpixels respectively.  ${\left\| C \right\|_{1,2}}$ is defined as $\sum\nolimits_{i = 1}^N {{{\left\| {{c^i}} \right\|}_2}}$, which is the sum of $l_2$ norms of rows in coefficient matrix $C$. Moreover, the constraint $diag(C) = {\bf{0}}$ forces the diagonal elements of matrix $C$ to be 0, which is to avoid  self-reconstruction.

After solving the above optimization problem, the obtained coefficient matrix $C$ is used to find the representative motion patterns. In detail, the $i^{th}$ row of matrix $C$ denoted as $c^i$, indicates the reconstruction coefficient of the $i^{th}$ motion feature in matrix $Y$. Therefore, the motion feature in matrix $Y$ whose corresponding reconstruction coefficient is nonzero has certain efforts to reconstruct other motion features and can be chosen as the representatives. Besides, the optimal coefficient matrix $C$ also provides information about ranking, $i.e.$, relative importance  of the representatives to describe the other normal superpixels' motion patterns. More precisely, a representative is essential to reconstruct many superpixels' motion patterns. Thus, its corresponding row in the optimal coefficient matrix $C$ contains many nonzero elements with large values. On the other hand, a representative only takes part in the reconstruction of few superpixels' motion pattern, hence, its corresponding row of $C$ contains a few nonzero elements with smaller values. Therefore, we rank $m$ representatives ${{\bf{y}}_{{i_1}}}, \ldots ,{{\bf{y}}_{{i_m}}}$ according to the relative importance, $i.e.$, ${\bf{y}}_{{i_1}}$ has the highest rank and ${\bf{y}}_{{i_m}}$ has the lowest rank. Whenever for the corresponding rows of $C$ we have
\begin{equation}
\label{Eq-Dict-select}
{\left\| {{c^{{i_1}}}} \right\|_2} \ge \left\| {{c^{{i_2}}}} \right\|_2 \ge  \cdots  \ge \left\| {{c^{{i_m}}}} \right\|_2.
\end{equation}
According to the ranking result, we select the top $M$ representatives to form the normal dictionary $D$, and the spatial localizations of the selected representatives denoted as $L$, are collected in the same order. Finally, the proposed optimization programs in Eq. \ref{Eq-DictLearning-1} can be written as
\begin{equation}
\label{Eq-DictLearning-2}
\mathop {\min }\limits_C \;{\lambda_1} {\left\| C \right\|_{1,2}} + \frac{1}{2}\left\| {Y - YC} \right\|_F^2\quad s.t.\;\;diag(C) = {\bf{0}}
\end{equation}
in practice.

\subsection{Spatial-aware reconstruction for abnormality measurement}
\label{Anomaly measurement}
 Denote the learned motion orientation dictionary as $D_o^t$ at time $t$. For a newly observed superpixel motion orientation feature $y_{oi}^t$, we first calculate the spatial distance between this superpixel and every element in the dictionary, and then select the top $K$ nearest elements to form a new spatial-near dictionary $D_{ol}^t$. To determine the motion orientation anomaly, the superpixel motion feature $y_{oi}^t$ is reconstructed by $D_{ol}^t$ and the reconstruction cost is viewed as anomaly degree of the examined superpixel.  To be specific, the anomaly is defined as:
\begin{equation}
\label{Eq-OAnomaly-Measurement}
a_{oi}^t = EMD(y_{oi}^t,\;\;{D_{ol}^t}{\alpha_{oi}^t}),
\end{equation}
where $a_{oi}^t$ is the anomaly degree of the  ${i^{th}}$ superpixel in the flow orientation field and $\alpha_{oi}^t$ is optimal solution of the following sparse reconstruction problem:
\begin{equation}
\label{Eq-OReconst}
{\alpha_{oi}^t} = \arg \;\mathop {\min }\limits_\alpha \;\left\| {{y_{oi}^t} - D_{ol}^t\alpha} \right\|_F^2 + {\lambda _2}{\left\| \alpha \right\|_1}.
\end{equation}
With the calculated $a_{oi}^t$ of each superpixel, we utilize the max-min normalizer to put $a_{oi}^t$ into the range of [0,1]. The anomaly degrees of all superpixels are gathered to construct a motion orientation anomaly map $S_t^O$ for the $t^{th}$ frame in motion orientation level.

As for motion magnitude anomaly measurement, since the video is captured on a moving vehicle, their demonstrated motion is relative.  This makes the abnormal motion magnitude might be very similar to the normal ones and utilizing the reconstruction strategy is unable to fulfill this task. In order to alleviate this problem, the abnormality of motion magnitude is measured by the difference between abnormal motion magnitude feature and elements of its spatial-near dictionary. Moreover, the highest weight is set to the nearest elements. In detail, suppose the $y_{mi}^t$ denotes the superpixel's motion magnitude feature and $D_{ml}^t$ denotes corresponding spatial-near dictionary, the anomaly degree of superpixel in motion magnitude field is calculated as follows:
\begin{equation}
\label{Eq-MAnomaly-Measurement}
a_{mi}^t = \frac{1}{K}\sum\limits_{j = 1}^K {w_{ij} \times EMD(} y_{mi}^t,{D_{mlj}^t}),
\end{equation}
where ${D_{mlj}^t}$ denotes the $j^{th}$ element of spatial-near dictionary and $w_{ij} = {e^{ - \left\| {z_{oi}^t - {l_{mlj}^t}} \right\|_2^2}}$ gives the nearest element the highest weight. Similarly, after the normalization operation, we gather all the anomaly degrees of superpixels to construct a motion magnitude map $S_t^M$. Besides, for easier combination and visualization of the following Bayesion integration, we harness max-min normalizer to put $S_t^O$ and $S_t^M$ into range $[0,1]$. The final anomaly map is generated by integrating these two maps and the integration strategy is described in section \ref{Bayesian integration}.

To alleviate the influence of dynamic scene, the dictionaries need to be updated. We incrementally cumulate the new normal superpixels' motion features $Y_{nor}$ and get the updated training set ${Y_{new}} = [{D_e}\;\;{Y_{nor}}]$, where $D_e$ is the old dictionary. The obtained $Y_{new}$ will subject to the dictionary learning procedure to obtain the updated dictionary every $T$ frame, as discussed in Section \ref{Dictionary learning}.

\begin{figure}
\centering
\includegraphics[width=.5\textwidth]{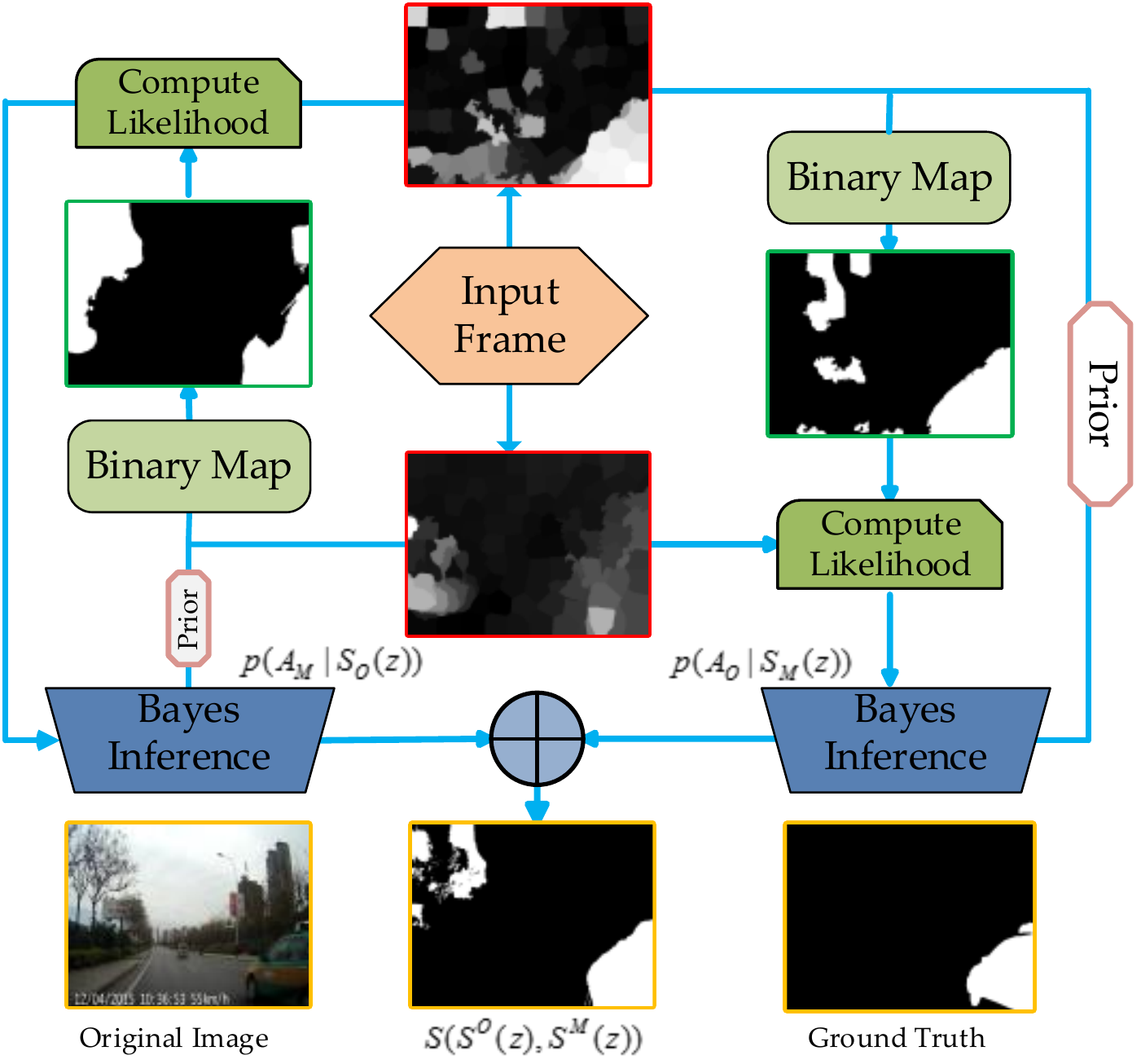}
\caption{Bayesian integration of anomaly maps The two anomaly maps are measured via motion orientation and magnitude respectively, denoted by $S_O$ and $S_M$.}\label{Fig-Bayes}
\end{figure}

\section{Bayesian-based integration of anomaly detection}
\label{Bayesian integration}
For anomaly detection in traffic scenes, the motion orientation and magnitude usually have different efforts in different cases, and are usually complementary to each other. Therefore, this work integrates the previously obtained two anomaly maps to generate the final anomaly map, which can address the change of vehicle velocity problem to some extent. To make full use of the complementarity between motion orientation and magnitude, this work employs an integration method based on Bayesian inference  \cite{DBLP:conf/iccv/LiLZRY13}. The posterior probability is formulated as:
\begin{equation}
\label{Eq-BayesionIntegration-model}
p(A|S(z)) = \frac{{p(S(z)|A)p(A)}}{{p(A)p(S(z)|A) + (1 - p(A))p(S(z)|N)}},
\end{equation}
where the prior probability $p(F)$ is a anomaly map, $A(z)$ is the anomaly degree of pixel $z$, $p(S(z)|A)$ and $p(S(z)|N)$ represent the detected abnormal and normal likelihood of pixel $z$, respectively. It is noted that the prior probability and the likelihood probabilities are the key points for the result.

Given  the motion orientation anomaly map $S^O$ and the motion magnitude anomaly map $S^M$, we treat one of them as the prior ${S^i}(i \in \{ M,O\} )$ and use the other one ${S^j}(i \ne j,j \in \{ M,O\} )$ to compute the likelihood, as shown in Figure \ref{Fig-Bayes}. Specifically, first,  $S^i$ is thresholded by its mean anomaly value and a binary map $B^i$ is obtained, the regions that having the value of 1 in binary map are denoted as $A_i$, which means abnormal regions. And the residual regions are normal regions, denoted as $N_i$.  In each region, the likelihood probability at pixel $z$ is calculated as:
\begin{equation}
\label{Eq-PLikelihood}
\begin{array}{l}
p({S^j}(z)|{A_i}) = \frac{{{N_{{A_i}b({s^j}(z))}}}}{{{N_{{A_i}}}}}\\
p({S^j}(z)|{N_i}) = \frac{{{N_{{N_i}b({s^j}(z))}}}}{{{N_{{N_i}}}}}
\end{array},
\end{equation}
where $N_{A_i}$ and $N_{N_i}$ are the number of the pixels in the detected abnormal region $A_i$ and the normal region $N_i$ in motion orientation map $S^i$. Moreover, the range [0,1] divides into $m$ intervals, and thus the $i-th, (i=1,2,...,m)$ interval is $[{{(i - 1)} \mathord{\left/{\vphantom {{(i - 1)} m}} \right.\kern-\nulldelimiterspace} m},{i \mathord{\left/ {\vphantom {i m}} \right.\kern-\nulldelimiterspace} m}]$. $b({s^j}(z))$ represents the interval where ${s^j}(z)$ falls into its range. ${N_{{A_i}b({S^j}(z))}}$ denotes the number of detected abnormal region's pixels whose value falls into $b({s^j}(z))$. Similarly, ${N_{{N_i}b({S^j}(z))}}$ represents the number of normal region's pixels whose values fall into $b({s^j}(z))$.

Consequently,  the posterior probability is computed with $S^i$ as the prior by
\begin{equation}
\label{Eq-Pposterior}
p({A_i}|{S^j}(z)) = \frac{{{S^i}(z)p({S^j}(z)|{A_i})}}{{{S^i}(z)p({S^j}(z)|{A_i}) + (1 - {S^i}(z))p({S^j}(z)|{N_i})}}.
\end{equation}
Similarly, we can also get $p({A_j}|{S^i}(z))$  by treating the two maps as the other. After obtaining the two posterior probabilities and specifying $i, j$ with $O, M$, we compute an integrated anomaly map $S(S^O(z),S^M(z))$, based on Bayesian integration:
\begin{equation}
\label{Eq-AnomalyIntegration}
S({S^O}(z),{S^M}(z)) = (p({A_O}|{S^M}(z)) + p({A_M}|{S^O}(z)))/2.
\end{equation}

The proposed Bayesian integration of anomaly maps is illustrated in Figure \ref{Fig-Bayes}. It should be noted that Bayesian integration serve these two maps as the prior in turn and cooperate with each other in an effective manner, which uniformly highlights abnormal objects in a frame.

\section{Experiments and discussion}
\label{experments and discussion}
In this section, we first introduce the datasets and implementation setups for the experiments. Then for demonstrating the effectiveness of the proposed method, we conduct experiments and compare the  results with other competitors. Finally,   analyses  and discussions are made to explain the  experimental results.
\subsection{Datasets}
\begin{figure*}
\centering
\includegraphics[width=0.95\textwidth]{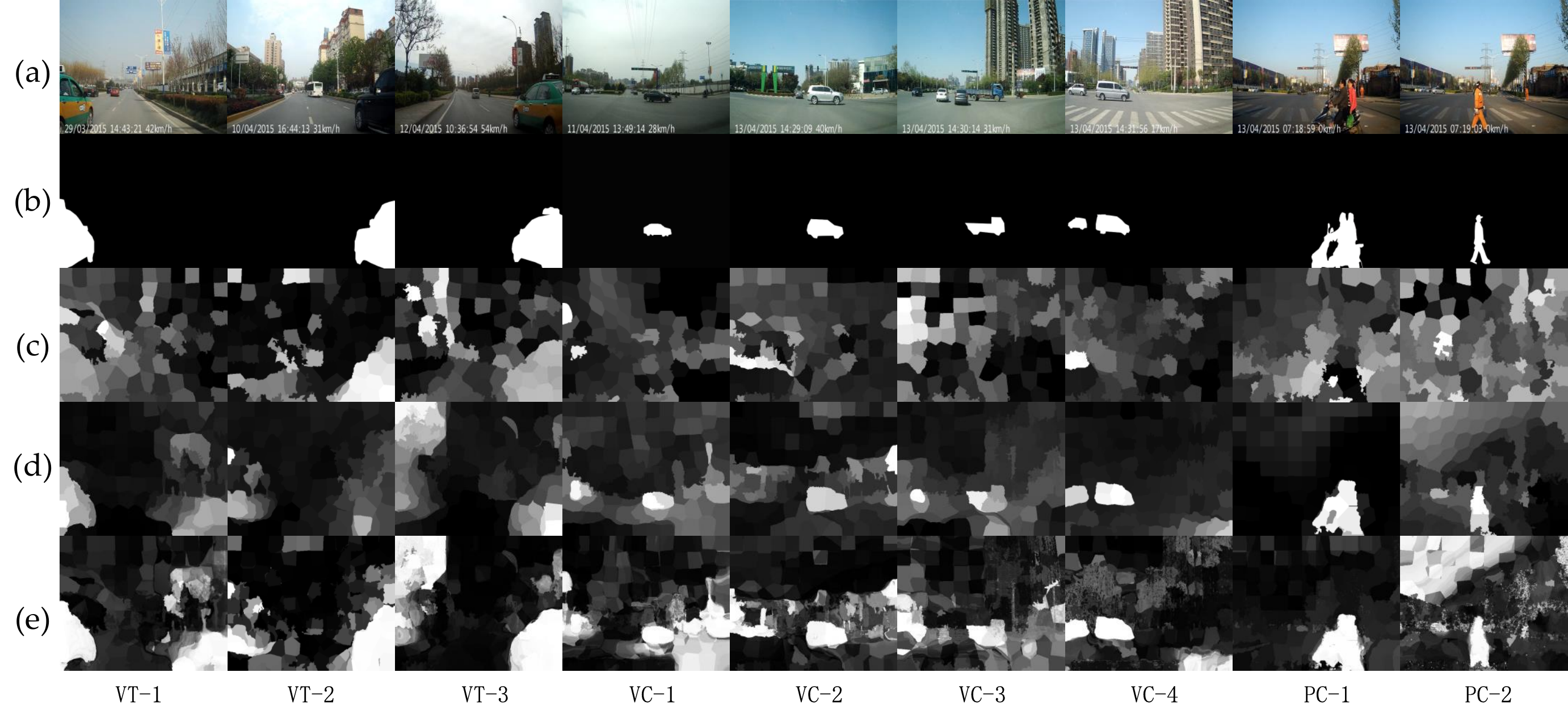}
\caption{Typical frameshots of the detected results by different competitors for each sequence. (a) Original color image; (b) Ground-truth anomaly; (c) Motion orientation anomaly map; (d) Motion magnitude anomaly map; (e) The integrated anomaly map.}\label{Fig-MapResult}
\end{figure*}

Since the publicly available datasets are almost captured by a static camera, such as the car accident\cite{5597797} dataset and QMUL Junction\cite{loy2009modelling} dataset, this paper provides a dataset consisted of nine driving videos, which contains several kinds of abnormal events. The videos are captured by a vehicle mounted camera for daily driving, and its view of angle is consistent with the driver's. The anomaly that we considered here is a kind of threats, which have potential dangers, such as vehicle overtaking. To be more specific, based on the anomaly types, the captured video sequences can be divided into three categories: 1) "Three sequences have the vehicle overtaking(VT) behavior (We name them as \emph{VT-1},\emph{VT-2}, and \emph{VT-3})", 2) "Four sequences consist of vehicle crossing(VC) behavior (They are named as \emph{VC-1}, \emph{VC-2}, \emph{VC-3} and \emph{VC-4})", 3) "Two sequences contain pedestrian crossing and motorcyclists crossing(PC) behaviors (They are denoted as \emph{PC-1} and \emph{PC-2})". Due to the online application of our method, we do not split the overall dataset into training and test part. And the first $10$ frames of sequences, which are always normal situation, are treated as training data for this sequence and the rest are utilized to test. There are $180$ frames in each sequence averagely, and the frameshots of the video sequences are demonstrated in Fig. \ref{Fig-MapResult}, some of which are very difficult for road anomaly detection because of complex background. In the captured dataset , the resolution of each frame is $480 \times 640$. The ground truth of each video sequence is manually labeled by ourselves.

\subsection{Implementation setup}
\subsubsection{Metrics}
In order to prove the efficiency of the proposed method, the qualitative and quantitative evaluations are both considered. For qualitative evaluation, we demonstrate several typical snapshots of the detected anomaly in each video sequence. As for the quantitative evaluation, pixel-wise receiver of characteristics (ROC) and area under ROC (AUC) are employed. Among them, ROC represents the detection ability of the proposed method, and its indexes are specified as:
\begin{equation}
\label{Eq-TPR-FPR}
TPR = TP/P,FPR = FP/N,
\end{equation}
where $TP$ denotes the number of the pixels truly detected, $FP$ is the number of pixels falsely detected, $P$ and $N$ represent the positive pixel number and negative pixel number, respectively.
\subsubsection{Parameters}
In our work, the SLIC superpixel\cite{achanta2012slic} is employed, in which $\delta$ represents the compactness and $N$ the number of the superpixels. The larger $\delta$ is, the more compact the superpixels are. In this paper, $\delta$ and $N$ are set as 10 and 125 for all sequences respectively. The dimension of the motion feature $c$ is specified as 30. Furthermore, the parameters $\lambda_1$ and $\lambda_2$ in Eq. \ref{Eq-DictLearning-2} and Eq. \ref{Eq-OReconst} are set as 0.5 and 0.5 in all experiments, respectively. The number of basis $M$ in the dictionary is set as 300 and the dictionary updating period $T$ is set as 5, which makes the dictionary over-complete all the time. Additionally, the size of spatial-near dictionary $K$ is set as 10.
\subsubsection{Comparisons}
\label{section-Comparisons}
Since the proposed method is fulfilled by the collaboration of the motion orientation and magnitude, the effectiveness of motion anomaly detection technique is firstly evaluated. In order to demonstrate the advantage of the proposed two-path motion description method, denoted as TPMD. we replace the proposed motion histograms with traditional histogram of oriented optical flow (THOOF) and measure the abnormality based on the proposed spatial-aware sparse reconstruction(SSRC-THOOF). Apart from spatial-near sparse reconstruction, we also make a comparison with other popular one class classification method. To be specific, we investigate one class SVM and Isolation Forest (IF) \cite{liu2008isolation}, which is a popular anomaly detection model based on random forest. These two variants are referred as SVM-THOOF and IF-THOOF respectively. Similarly, we retain TPMD and replace the proposed spatial-aware reconstruction method with traditional sparse reconstruction (SRC) \cite{DBLP:journals/pr/CongYL13}, one class SVM and Isolation Forest (IF) ,which are denoted by SRC-TPMD, SVM-TPMD and IF-TPMD respectively. It should be noted that these three variants do not take the spatial information into consideration. Finally, we refer to our method as SSRC-TPMD and make a comparison between performances the proposed method and the above two variants and do some analysis according to the results.

As the second part of the proposed method, we integrate the two aspects to get the final result. To further validate the proposed integration method, we compare the detection results without integration and with different integration methods. To be specific, the competitors are motion Magnitude (M) detection result, motion Orientation(O) detection result, integration result using inner-product of motion magnitude and motion orientation (MO), integration result using our Bayes model(B-MO).

Last but not the least, for demonstrating the superiority of our method, it is in comparison with recent region proposal-based object detector Faster-RCNN \cite{ren15fasterrcnn}, which outperforms significantly traditional object detection method. And it is noted that region proposal-based object detection technique can boost our system and achieve a higher performance.

\begin{table*}[htbp]
\renewcommand\arraystretch{1.3}
\centering\caption{The AUC(\%) comparison of different descriptor and classification method. For a clear and fairer comparison. The bold one is the best result.}
\begin{tabular}{p{1.5cm}p{1cm}p{1cm}p{1cm}p{1cm}p{.2cm}p{1cm}p{1cm}p{1cm}p{1cm}}
\toprule[1.5pt]
\multirow{2}*{\textbf{Category}}  &  \multicolumn{9}{c}{\textbf{Method}} \\ \cline{2-10}
                                         & SVM-THOOF & IF-THOOF &SRC-THOOF & SSRC-THOOF & & SVM-TPMD & IF-TPMD & SRC-TPMD  & SSRC-TPMD(our) \\ \cline{2-5} \cline{7-10}
VT &65.94&69.96&72.75&81.89&   &70.88&78.91&68.47 &\textbf{85.38} \\
VC &80.96 &75.13 &83.64 &81.43 &   &79.73 &\textbf{89.23}&83.54 &88.44 \\
PC &78.16 &87.54 &82.75 &88.65 &   &74.47 &\textbf{92.27} &91.16 &90.61 \\ \hline
Average &75.34 &76.16 &79.80 &83.19 &   &75.61 &86.46 &80.21 &\textbf{87.90} \\
\bottomrule[1.5pt]
\end{tabular}\label{Table-compare-variants}
\end{table*}

\subsection{Evaluation of motion anomaly detection}
\subsubsection{Descriptor Comparison}
\begin{figure}[!h]
\centering
\includegraphics[width=.45\textwidth]{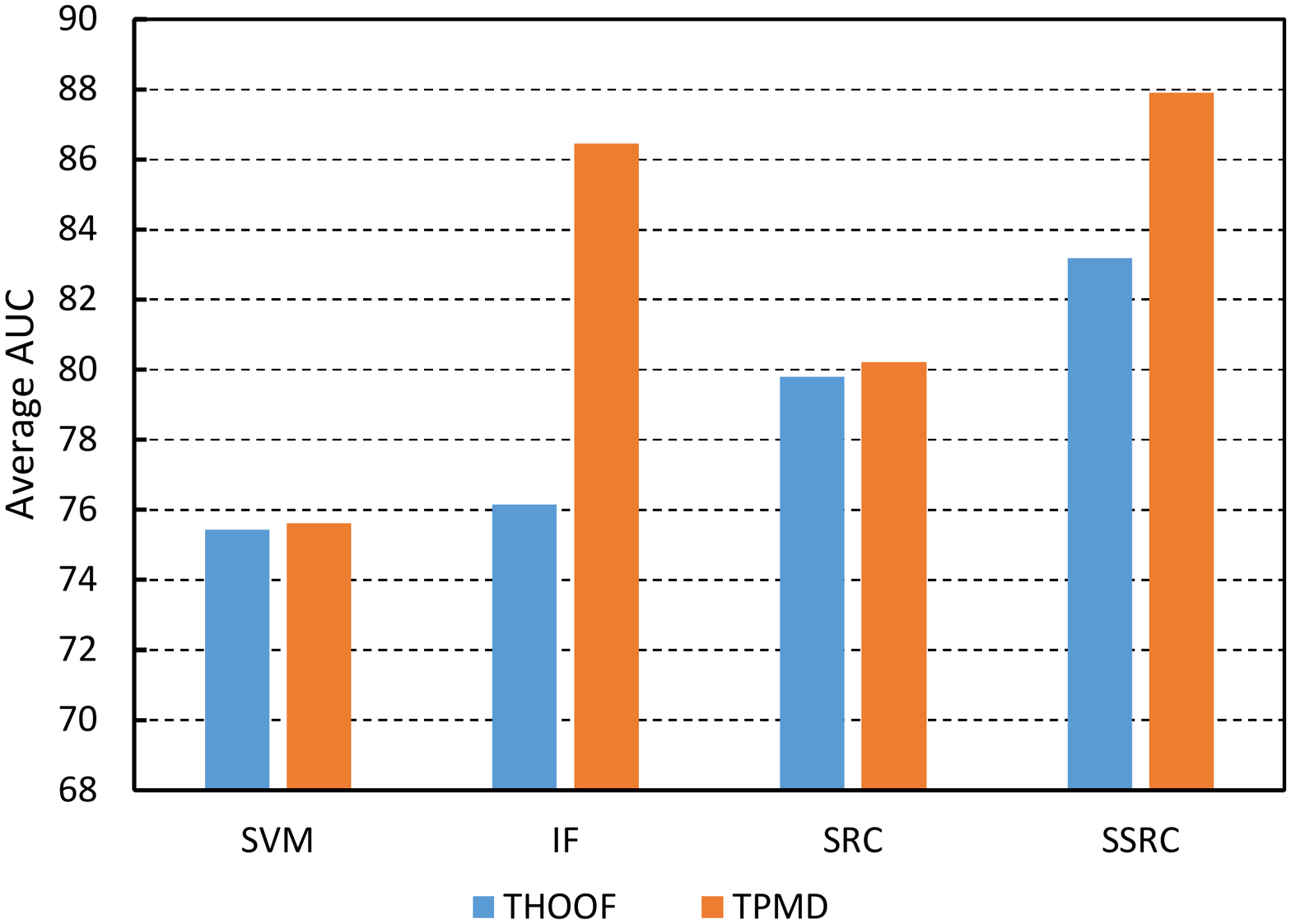}
\caption{The AUC value comparison of SVM, IF, SRC \cite{DBLP:journals/pr/CongYL13} and our method for each sequence.}\label{Fig-compare-descriptor}
\end{figure}

The first experiment evaluates the benefits of the two-path motion descriptor(TPMD). The THOOF descriptor \cite{chaudhry2009histograms} is proposed to describe motion characteristic of sequences, and it pours attention into motion direction information \cite{Yang2011Sparse}. Our motion utilization strategy is inspired by THOOF, and compute another histogram to describe motion energy information precisely. Therefore, in order to justify the superiority of the proposed TPMD, we combine these two descriptors with serval popular classifier, and average AUC values for each behavior category are listed in Table \ref{Table-compare-variants}. For a better visualized comparison, Fig. \ref{Fig-compare-descriptor} illustrates the difference between THOOF and TPMD, and the performance of every method is represented by the average AUC value over the total nine sequences. It can be seen that every TPMD-based method outperforms the corresponding THOOF-based variants, and there are 3.9\% improvement medially. From this performance comparison, the superiority of our TPMD is apparently verified.

\subsubsection{Classifier Comparison}
\begin{figure*}[!t]
	\centering
	\subfigure[The performance of THOOF-based variants]{
		 \label{SubFig-THOOF}
		\includegraphics[width=0.4\textwidth]{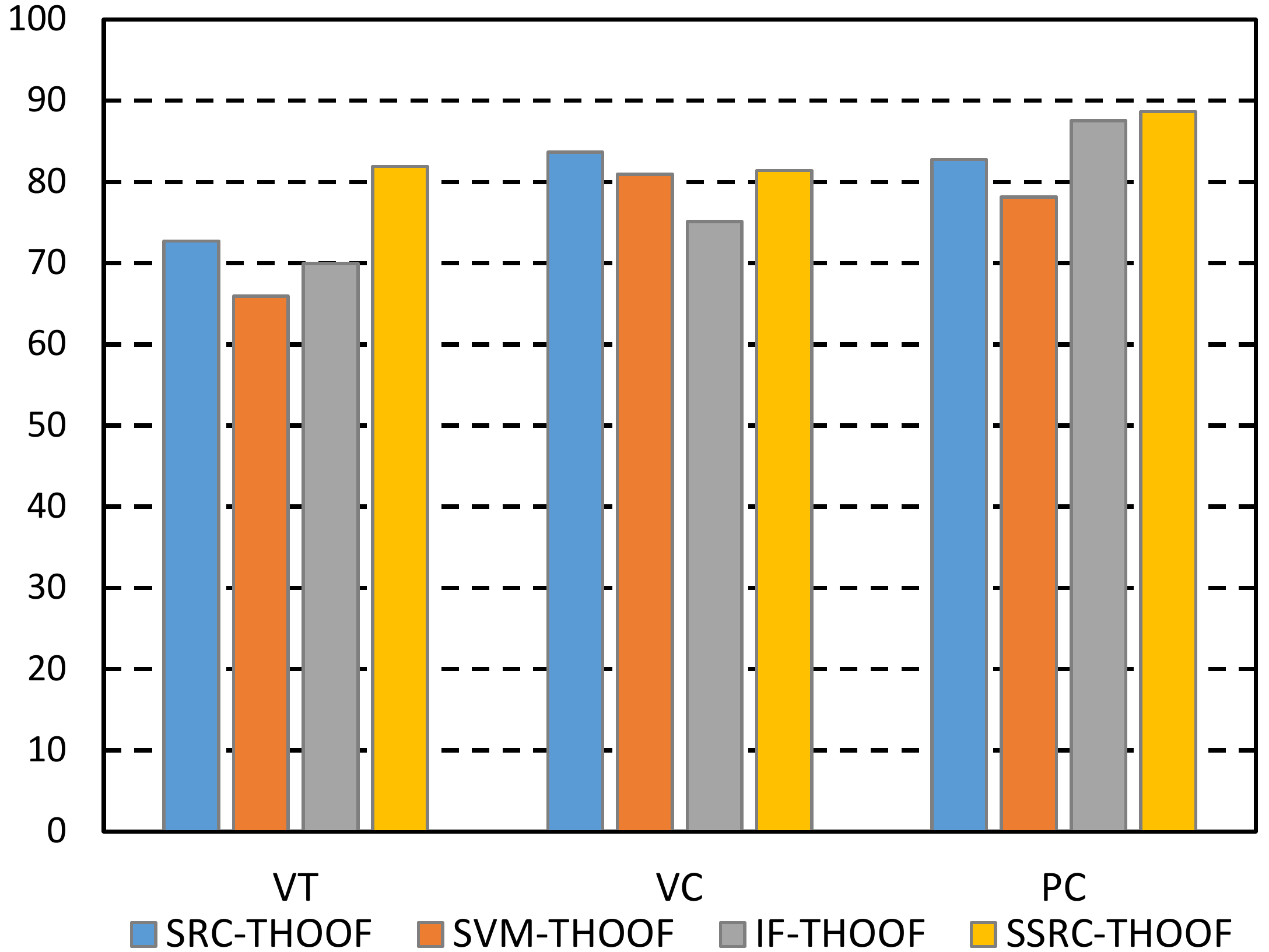}}
	\subfigure[The performance of TPMD-based variants]{
		\label{SubFig-TPMD}
		\includegraphics[width=0.4\textwidth]{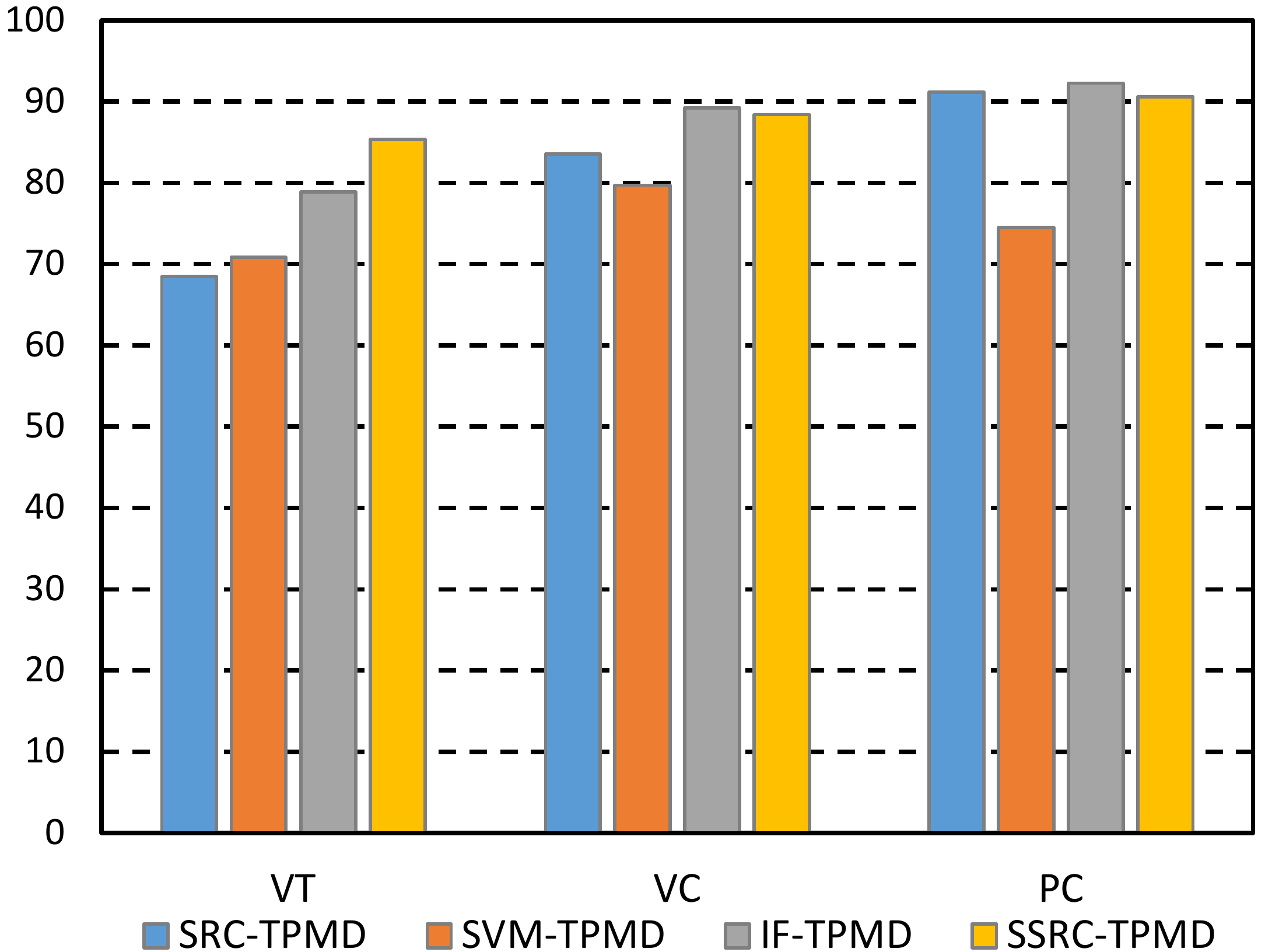}}
	\caption{The performance of all competitors, for detailed comparison, the average AUC value of every category is plotted.} \label{Fig-compare-spatail}
\end{figure*}

We next investigated the advantage of the spatial-aware sparse reconstruction, with the Bayesian integration method. Apart from sparse reconstruction, there are a slice of wide-used classification methods, such as Support Vector Machine(SVM), Artificial Neural Network(ANN) and Random Forest(RF). Because ANN is usually utilized to classify two or more classes, and it is not suited for one class problem, in addation to traditional traditional sparse reconstruction (SRC) \cite{DBLP:journals/pr/CongYL13}, the SVM and RF are selected as competitors. In detail, traditional SRC ,one-class SVM in \cite{chang2011libsvm} and Isolation Forest (IF) \cite{liu2008isolation} replace the sparse-aware reconstruction, and other parts stay the same. IF explores the concept of isolation with random forest for anomaly detection and achieves pleasurable performances in many application. It should be noted that these three competitors do not take spatial information into consideration. The performance of overall dataset is summarized in Fig. \ref{Fig-compare-spatail}. From the shown results, our method generates favorable accuracy for every behavior category regardless of the adopted desccriptor. To be specific, in the right sub-figure in Fig. \ref{Fig-compare-spatail}, our method performs best for VT behavior, and is comparable with best performer for other behaviors. A strong competitor is IF-TPMD and generates superior results in serval sequences, but it is not robust to anomaly type. In particular, the IF classification method performs worse than our method in detecting vehicle overtaking, while our method is independent of specific events. Moreover, because our spatial-aware sparse reconstruction makes modification to traditional SRC, we conduct a comparison between SRC and SSRC. As shown in Fig. \ref{Fig-compare-spatail}, SSRC significantly outperforms SRC in almost every behavior(improvement of AUC by as much as 7 percent), and this suggests the spatial information is crucial to higher accuracy.

Similar conclusion comes with the left sub-figure in Fig. \ref{Fig-compare-spatail}, where the THOOF is treated as motion descriptor. In addition, the SVM-based variants perform worse than others whatever the descriptor was used. This is not totally surprising, given the instability of optical flow. In other words, the noise of optical flow, which is caused by camera motion and dynamic background, makes SVM ineffective in this case. That is to say, our method can eliminate the influence of noise.

\begin{table}[htbp]
\centering
\caption{The AUC(\%) comparison of different clue and integration method. For a clear and fairer comparison. The bold one is the best result.}

\begin{tabular}{|c|c|c|c|c|c|}
\hline
 Sequence &Superpixels &O &M &MO &B-MO \\ \hline \hline
 VT-1 &125 &85.02 &77.33 &\textbf{85.16} &81.68 \\ \hline
 VT-2 &125 &76.07 &81.79 &82.17 &\textbf{87.56} \\ \hline
 VT-3 &125 &\textbf{89.03} &76.97 &86.53 &86.91 \\ \hline
 VC-1 &125 &50.97 &81.67 &59.89 &\textbf{89.68} \\ \hline
 VC-2 &125 &53.16 &\textbf{90.03} &80.73 &89.18 \\ \hline
 VC-3 &125 &39.20 &80.64 &63.59 &\textbf{83.33} \\ \hline
 VC-4 &125 &69.96 &89.83 &90.06 &\textbf{91.54} \\ \hline
 PC-1 &125 &57.11 &83.96 &73.10 &\textbf{87.33} \\ \hline
 PC-2 &125 &54.55 &84.55 &78.81 &\textbf{93.89} \\ \hline
 Average &- &63.89 &82.97 &77.78 &\textbf{87.90} \\ \hline
\end{tabular}\label{Table-AUC}

\end{table}

\subsection{Evaluation of integration method}
To further explore the effectiveness of the Bayesian integrated model, the performance comparisons are presented in Table \ref{Table-AUC}. It can be seen that the Bayesian integration model is superior to the other integration techniques. In addition, we also make a comparison between motion magnitude (M) and motion orientation (O) anomaly detection result, and it is noticed that motion magnitude and orientation have different importance in different sequences. Specifically, for the sequences containing vehicle overtaking behavior, ($i.e.$, VT-1, VT-2 and VT-3,) the motion orientation anomaly detection result is usually superior to the motion magnitude anomaly detection result, $i.e.$, $O>M$. The reason is that the motion orientation of abnormal object is very different from the background or normal object. However, the motion magnitude may be very similar to background. But on the other hand, the motion magnitude anomaly detection result has a higher performance in other sequences. The reason is that motion magnitude of abnormal object is very different from background or normal object, but the motion orientation not. The above phenomenon is caused by the different relative speed between the abnormal object and the mobile camera. Generally, the overtaking vehicle usually has a faster speed than the camera. Therefore, the estimated optical flow can represent motion magnitude and motion orientation well. However, for vehicle crossing behavior, the crossing objects usually have a slower speed than the camera. Therefore, the estimated optical flow can only represent motion magnitude well, as illustrated in Fig. \ref{Fig-motionShow}. Besides, it is noticed that the motion magnitude anomaly detection result  has a high performance in all sequences, and it demonstrates the proposed motion magnitude descriptor is effective. In general, the method using only motion magnitude or motion orientation can not handle all kinds of abnormal events in traffic scenes because of the different relative speeds between abnormal object and the camera. In order to make use of these two aspects simultaneously, this work reasonably integrates both detection results.

For this purpose, this work integrates the motion magnitude and orientation detection results based on a Bayesian model. In order to demonstrate its effectiveness of the Bayesian model, we make a comparison between Bayesian integration (B-MO) and naive integration technique (MO), which is achieved by making inner-product using both aspects. It is manifest in Table \ref{Table-AUC} that the performance of MO sometimes is lower than M or O (for example, VC-1, VC-2, VC-3). This implies that the naive integration technique can not boost the performance, but weaken it. The reason is that a high performance using inner-product needs high performances in both aspects, but it is impossible for some sequences to get satisfying results in both aspects. In order to make use of their complementarity, this work integrates both detection results based on Bayesian model. From Table \ref{Table-AUC}, it can be seen that the performances based on Bayesian model are almost the highest in most sequences.  Therefore, the integration technique can generate a high performance even though one single aspect  has a very low performance. According to the above analysis, we can conclude that the Bayesian integration model is better than the naive method.

\begin{table}[htbp]
\renewcommand\arraystretch{1.3}
\centering\caption{The AUC(\%) comparison of SSRC-TPMD and Faster-RCNN. For a clear and fairer comparison. The bold one is the average value.}
\begin{tabular}{c|c}
\toprule[1.5pt]
 &\textbf{AVERAGE} \\
Faster-RCNN &\textbf{86.15} \\
SSRC-TPMD &\textbf{87.90} \\ \hline
Faster-RCNN+SSRC-TPMD &\textbf{93.11} \\
\bottomrule[1.5pt]
\end{tabular}\label{Table-Faster-RCNN}
\end{table}

\subsection{Performance comparison}
Recently, region proposal technique achieves a great success in detecting objects from a image and is adopted in different works such as Markus Enzweiler's pedestrian detector \cite{enzweiler2009monocular},  Will Zou's work on regionlets \cite{wang2013regionlets}, etc. For demonstrating the superiority of our method, region proposal-based object detector is regarded as competitor. Specifically, the Faster-RCNN is tested on dataset and its performances are listed in Table \ref{Table-Faster-RCNN}. There are several reasons behind selecting Faster-RCNN as competitor, in the first place, the CNN-based Faster R-CNN achieves state-of-the-art performances on almost all public object detection datasets and outperforms Markus Enzweiler's pedestrian detector as well as Will Zou's work on regionlets. There is one more point, I should touch on, that traditional object detection methods are very dependent on specific dataset and is difficult to transfer to another dataset. Therefore, because of insufficient training data of our dataset, Faster-RCNN is our best choice. The last but not the least, albeit we do not fine-tune Faster-RCNN with our own data, the pre-trained model is robust to changing scenes and generates a promising results in our dataset. As shown in Table \ref{Table-Faster-RCNN}, the performance of Faster-RCNN is inferior to our method with only 1 percent, and superiority of our algorithm is demonstrated.

Furthermore, because only appearance information is processed in Faster-RCNN, it is beneficial to incorporate it into our method, which just utilizes the motion information. From Table \ref{Table-Faster-RCNN}, there is a significantly improvement after incorporation. As for incorporating strategy, we just add the object detection score on anomaly map and re-normalize it into range $[0,1]$.

\subsection{Discussions}
\subsubsection{Range of moving objects' speed}The motion estimation in our algorithm is highly dependent on the object's speed, and one basic assumption behind optical flow method is that object's movement is small between continuous two frames. Therefore, it is important to specify the range of moving objects' speed. Because we can not estimate the speeds of all objects in the scene accurately, we just record the speed of the camera. Moreover, there is two point that can explain the rationality of replacing objects' speed with camera's. First, the moving objects is quite fewer than static objects in the video frames, and their speeds in the video is just camera speed. Moreover, the moving objects's speeds in the videos are usually lower than static objects', and the reason behind this that objects almost are moving in the same direction as camera. Therefore, camera speed usually represents the highest speed in the frame and can be used to specify the range of moving objects' speed. Another important reason behind the collection of camera speed is that the absolute speeds of objects are useless. In detail, due to the mobile camera, the object moving speed in captured video is relative speed. For example, the static building is moving at $40km/h$ in video when camera speed is $40km/h$. Therefore, instead of specifications of the range of moving objects' speed, camera speed, which is obtained according to the vehicle's speedometer, is recorded to explain the system's robustness to motion speed. Numerically, the camera speed varies from $0km/h$ to $60km/h$ in dataset video, which almost cover the highest speed limitation in urban road, and there is no problem with our system in this speed range. The effectiveness of our method with higher camera speed is not probed now, and a deeper investigation will be done in the future.

\subsubsection{Runtime}In this paper, our method is achieved by a MATLAB-implementation on a machine with Intel i5-3470 3.2GHz CPU and 4GB RAM. The main consumption is taken by SLIC superpixel segmentation whose average runtime at 125 superpixels is about 0.353s. The spatial-aware reconstruction is very fast and only costs 0.083s, and Bayesian integration of two anomaly maps takes away 0.169s. Despite our work requires computing two anomaly maps, the superpixel segmentation and spatial-aware reconstruction are running in parallel, and will not double the time. Therefore, the total average runtime of this work is 0.605s without code optimization. Albeit our method cannot achieve real-time speed, it is faster than many pedestrian detectors, such as ChnFtrs(0.845s)\cite{DollarBMVC09ChnFtrs}, LatSvm-V2(1.589s)\cite{lsvm-pami} and MultiFtr+CSS(37s)\cite{437}. For a real-time consideration, we will use some accelerating strategy to make the method perform in real-time.


\section{Conclusion}
\label{conclusion}
This work addresses the problem of anomaly detection in traffic scenes from a driver's perspective, which is important to autonomous vehicles in intelligent transportation systems. In order to tackle three main difficulties caused by the mobile camera, this work describes motion magnitude and orientation respectively, and by measuring the abnormality of these two aspects simultaneously in conjunction with an adaptively weighted integration, the proposed method can alleviate the influence of the ever-changing scene and camera movement. Specifically, a new motion descriptor is presented to represent the motion magnitude and orientation by calculating a histogram respectively. It performs better than THOOF, which only describes the motion orientation information. With this motion descriptor, the motion anomaly is measured by the reconstruction cost of the spatial-near dictionary, and then these two clues are integrated by a Bayesian model to get a robust result. From the experimental results, the effectiveness and efficiency of the proposed method are proved. Some conclusions can be summarized through this work: 1) For describing the motion information more effectively, the calculated two motion histograms can describe motion magnitude and motion orientation respectively, and it is better than the THOOF. 2) Compared with the traditional anomaly detection, the spatial locations of motion patterns play an essential role in traffic scene anomaly detection. In order to utilize this spatial location information, this work measures the abnormality of the motion orientation and magnitude by reconstructing it over its spatial-near dictionary, and the experimental results demonstrates the rationality of the proposed method. Moreover, the influence of dynamic background is eliminated to some extent. 3) With the obtained two motion anomaly maps, this work fuses them based on a Bayesion-based integration method, which makes use of the complementarity of the two anomaly maps and the obtained result is robust to the change of vehicle velocity.

In the future, we would like to use more clues, for example, near-infrared information, depth information and so on, to improve the performance and robustness of the proposed method. Based on these new information, we would like to extend our method to handle more kinds of abnormal events. The key point is how to use these clues reasonably and integrate them efficiently.


\ifCLASSOPTIONcaptionsoff
  \newpage
\fi

\bibliographystyle{IEEEtran}
\bibliography{IEEEabrv,itspaper}

\begin{IEEEbiographynophoto}{Yuan Yuan} (M'05-SM'09) is currently a full professor with the School of Computer Science and Center for OPTical IMagery Analysis and Learning (OPTIMAL), Northwestern Polytechnical University, Xi'an 710072, Shaanxi, P. R. China. She has authored or coauthored over 150 papers, including about 100 in reputable journals such as IEEE Transactions and Pattern Recognition, as well as conference papers in CVPR, BMVC, ICIP, and ICASSP. Her current research interests include visual information processing and image/video content analysis.
\end{IEEEbiographynophoto}

\begin{IEEEbiography}[{\includegraphics[width=1in,height=1.25in,clip,keepaspectratio]{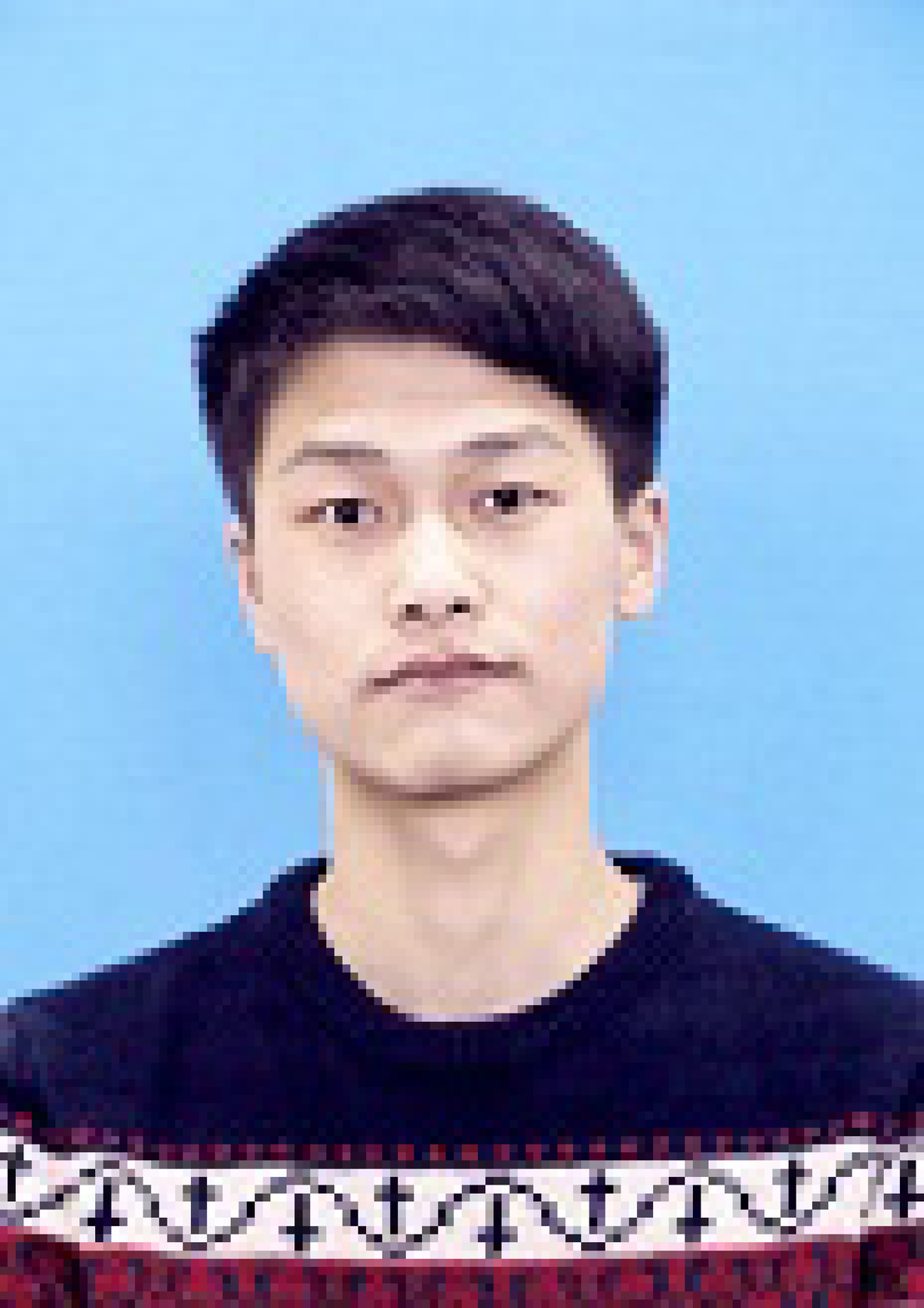}}]{Dong Wang} received the B.E. degree in Computer Sciense and Technology from the Northwestern Polytechnical University, Xi'an 710072, Shaanxi, P. R. China, in 2015. He is currently pursuing the Ph.D. degree from the Center for Optical Imagery Analysis and Learning, Northwestern Polytechnical University, Xi¡¯an, China. His research interests include computer vision and pattern recognition.
\end{IEEEbiography}

\begin{IEEEbiography}[{\includegraphics[width=1in,height=1.25in,clip,keepaspectratio]{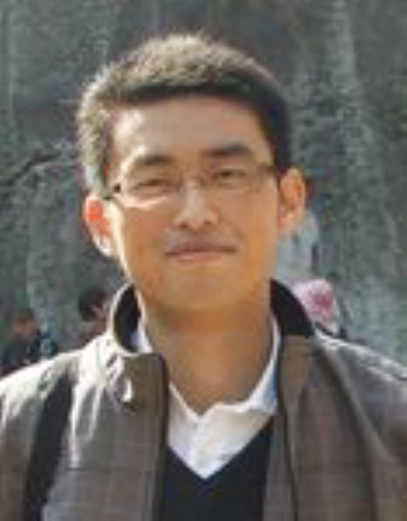}}]{Qi Wang} (M'15-SM'15) received the B.E. degree in automation and Ph.D. degree in pattern recognition and intelligent system from the University of Science and Technology of China, Hefei, China, in 2005 and 2010 respectively. He is currently an Associate Professor with the School of Computer Science and the Center for Optical Imagery Analysis and Learning (OPTIMAL), Northwestern Polytechnical University, Xi'an 710072, Shaanxi, P. R. China. His research interests include computer vision and pattern recognition.
\end{IEEEbiography}


\end{document}